# Feature Sharing Cooperative Network for Semantic Segmentation

Ryota Ikedo, Kazuhiro Hotta
*Meijo University, Japan*
160442012@ccalumni.meijo-u.ac.jp, kazuhotta@meijo-u.ac.jp

Abstract: In recent years, deep neural networks have achieved high accuracy in the field of image recognition. By inspired from human learning method, we propose a semantic segmentation method using cooperative learning which shares the information resembling a group learning. We use two same networks and paths for sending feature maps between two networks. Two networks are trained simultaneously. By sharing feature maps, one of two networks can obtain the information that cannot be obtained by a single network. In addition, in order to enhance the degree of cooperation, we propose two kinds of methods that connect only the same layer and multiple layers. We evaluated our proposed idea on two kinds of networks. One is Dual Attention Network (DANet) and the other one is DeepLabv3+. The proposed method achieved better segmentation accuracy than the conventional single network and ensemble of networks.

## 1 INTRODUCTION

Convolutional Neural Network (CNN) (Krizhevsky, A., 2012) achieved high accuracy in various kinds of image recognition problems such as image classification (Szegedy, C., 2015), object detection (Redmon, J., 2016), pose estimation (Cao, Z., 2018) etc. In addition, semantic segmentation assigns class labels to all pixels in an input image. This task recognizes various classes at pixel level. Semantic segmentation using CNN is also applied to cartography (Isola, P., 2017), automatic driving (Chen, L.C., 2018), medicine and cell biology (Havaei, M., 2017). CNN obtained high level feature by aggregating local features. In recent years, some works based on Fully Convolutional Networks (FCN) (Long, J., 2015) had been proposed to enhance feature representations. One of the methods used multi-scale context fusion (Ding, H., 2018). Some works (Yang, M., 2018) aggregated multi-scale contexts by using dilated convolutions and pooling. Some papers (Peng, C., 2017) extracted richer global context information by introducing effective decoder to the network. In other cases, there are some works (Huang, Z., 2019) using attention mechanisms for semantic segmentation. As described above, many researchers aim to improve the segmentation accuracy by various methods. Here we propose new learning method for semantic segmentation.

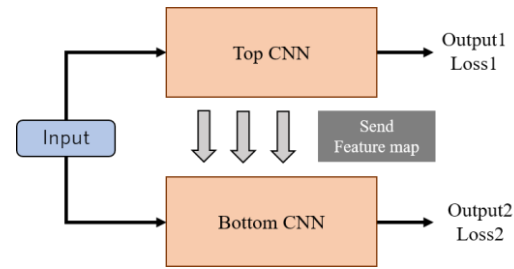

Figure 1. The structure of Cooperative network

In general, the weights of CNN changes depending on the initial values of a random function even if we use the same network. The feature maps obtained from the network also change inevitably. We focused on the difference between networks. We want to improve the quality of learning by creating a cooperative relationship between feature maps from each network. We call this network "cooperative network". The overview of our proposed method is shown in Figure 1. At first, we prepare two same CNNs and train both CNNs simultaneously. Then we introduce some paths between two networks, and the feature maps of top network are sent to the bottom network. This path allows each network to learn while sharing the information. In this way, it is possible to perform cooperative learning that incorporates the beneficial different information while sharing information.

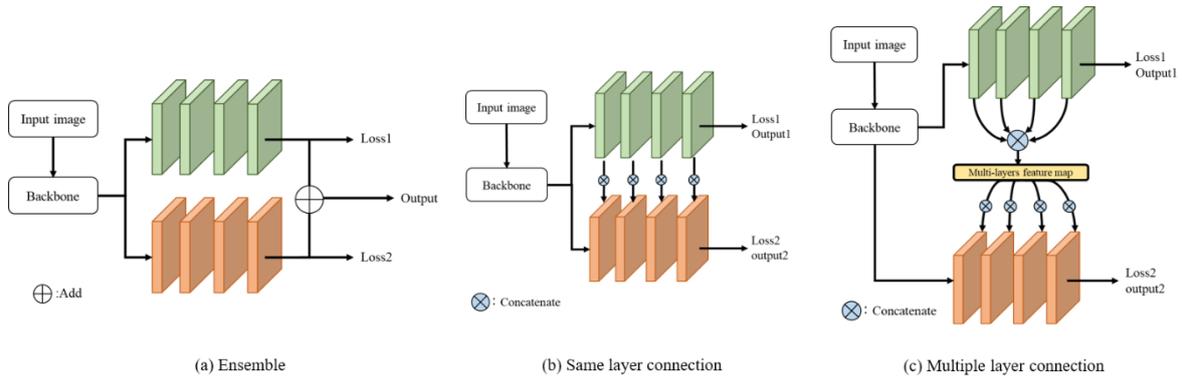

Figure 2. The structure of ensemble network (a) and the proposed cooperative network (b)(c)

We conducted experiments on two famous networks to confirm the effectiveness of our method. First, we use Dual Attention Network (DANet) (Fu, J., 2019) which introduces two kinds of attention mechanisms. Second, we use DeepLabv3+ (Chen, L.C., 2018) which applies the depthwise separable convolution to both atrous spatial pyramid pooling (ASPP) (Zhang, H., 2018) and decoder modules. In experiments, we evaluate the proposed method on the Cityscapes dataset (Cordts, M., 2016) and PASCAL VOC2012 dataset (Everingham, M., 2010). We confirmed that our proposed cooperative network gave higher accuracy than conventional single network and the ensemble of networks in both experiments.

This paper is organized as follows. In section 2, we describe related works. The details of the proposed method are explained in section 3. In section 4, we evaluate our proposed cooperative learning on segmentation tasks. Finally, we describe conclusion in section 5.

## 2 RELATED WORK

The state-of-the-art approaches for semantic segmentation are mainly based on CNNs. The famous approach is based on FCN structure such as SegNet (Badrinarayanan, V., 2017), U-net (Ronneberger, O., 2015) and so on. They are the simple structure of FCN. In recent years, sharp accuracy improvements have been driven by new architectures. One of the problems in semantic segmentation is that CNN lost spatial information by reducing the resolution in feature extraction processes. Dilated convolution was proposed to solve this problem. It can extract features while preserving spatial information by expanding receptive fields sparsely without reducing resolution.

In the other works, PSPNet (Zhao, H., 2017) and DeepLab (Chen, 2018) proposed ASPP module. ASPP module aggregates feature information at multiple scales. These works can get multi-scale contextual information and achieved high accuracy. Recent some studies adopted attention module (Takikawa, T., 2019) for segmentation task. For example, DANet introduced two kinds of attention modules which capture contextual information in spatial and channel domains.

U-net used the connections between encoder and decoder with the same resolution. The connections send the information of fine objects and correct position of objects from encoder to decoder, and the information helps to improve the segmentation accuracy. This is the cooperation between encoder and decoder. In this paper, we consider the cooperation between multiple CNNs in order to improve the segmentation accuracy further.

Cooperation of networks was used in some tasks. For example, collaboration learning provides supplementary information and regularization to each classifier (Song, G., 2018). Thus, it introduced collaborative learning methods to help normalize and improve robustness without additional inference. In works (Zheng, H., 2018), collaborative network is used to fuse two kinds of data information. These works introduced collaborative learning to improve the accuracy by using multiple data. From these studies, we focus on different information which each network has. Therefore, we introduce cooperative learning method for sharing information that cannot be obtained by a single network.

# 3 PROPOSED METHOD

This section describes the details of the proposed method. We explain the details of networks in section 3.1. Proposed method 1 which connects only the same layers in two networks is explained in section 3.2. We explain the proposed method 2 which connects different layers in section 3.3.

## 3.1 Overview

Deep neural network is inspired from neuron connection of human brain. There are several approaches that they imitate learning methods of human for improving accuracy (Lake, B. M., 2015). We propose cooperative learning for deep neural network like people who study while exchanging knowledge among people. In our work, we train two CNNs at same time while interacting between them.

We show the structure of cooperative learning in Figure 1. If we prepare two networks and introduce the connections between two networks, the feature map obtained from the top network can be sent to the bottom network, and two feature maps are concatenated. Since both networks train to solve segmentation problem, good information for addressing the task is sent to the bottom network. Thus, the bottom network can use good feature maps obtained from the top network for solving task, and the bottom network focus on the problem that the top network cannot solve. Since bottom network concatenates the feature map obtained from top network, the number of filters in the concatenated feature map is two time larger than those of the original network. This is the cooperative learning that we propose in this paper.

In our method, we adopt the following loss because our method trains two CNNs separately at the same time. We use Softmax Cross Entropy for calculating the loss.

$$Loss = Loss1 + Loss2, \qquad (1)$$

where Loss1 is the loss for CNN1 and Loss2 is for CNN2. Both losses are optimized simultaneously. In this network architecture, we can train two networks while sharing feature maps that single network cannot obtain. There are several advantages of cooperative learning. First, we can increase the amount of useful information by using feature maps between two networks. Since the first network solves the segmentation task, the features for solving the task are already obtained. Thus, the second network can learn the task using the information from the first network as a reference. For the above reasons, we consider that cooperative learning is effective for improving accuracy.

In this paper, we propose two connection methods for cooperative learning. The first connection method is between same layers. The second connection method is between multiple layers. These two methods are explained in the following subsections. Furthermore, we confirm the effectiveness of cooperative connection by comparing with ensemble network. Ensemble network used two same CNNs. The difference between ensemble network and our method is shown in Figure 2 (a).

## 3.2 Same Layer Cooperative Network

We introduce the connection between the corresponding layers in two same networks. A structure of the proposed connection is shown in Figure 2 (b). This connection sends feature map in top network to the same layer in bottom network. We call this "same layer connection". Since two same networks are used, the same layer connection is the basic connection of the proposed cooperative learning. Since the loss for top network is optimized, the feature maps in top network are effective for solving the task. The bottom network can obtain those features for solving the task. Thus, it is expected that the accuracy will improve.

## 3.3 Multiple layer Cooperative Network

We also propose multiple layer connection method to give the information obtained at multiple layers in top network to the bottom network. In general, each layer in CNN has different kinds of information such as correct location and semantic information. We give those various kinds of information in top network to each layer in bottom network because information in different layers in top network may be effective to solve segmentation problem. We call this "multiple layer connection". Figure 2 (c) shows the structure.

We give the feature maps from the shallow to deep layer in top network to certain layer in bottom network to use both fine and semantic information well. We extract feature maps from top network and aggregate those feature maps. By sending the aggregating feature maps, bottom network obtains the information that single network and same layer connection cannot obtain, and it is useful to solve the task. However, the size of feature maps of multiple layers is different. Thus, we interpolated the feature maps to be the same size.

Table1.    Segmentation results on Cityscapes dataset

| Method | | Mean IoU | road | sidewalk | building | wall | fence | pole | traffic light | traffic sign | vegetation | terrain | sky | person | rider | car | truck | bus | train | motorcycle | bicycle |
|---|---|---|---|---|---|---|---|---|---|---|---|---|---|---|---|---|---|---|---|---|---|
| DANet | Single | 75.9 | 97.9 | 83.4 | 92.4 | **58.6** | 59.3 | 66.6 | 71.9 | 80.1 | 92.5 | 58.6 | 94.8 | 83.0 | 62.8 | 94.8 | 80.0 | 80.6 | 44.0 | 62.5 | 77.6 |
| | Ensemble | 76.8 | 97.5 | 81.4 | 91.9 | 58.2 | 57.2 | 65.6 | 71.1 | 79.7 | 92.6 | 61.0 | 94.6 | 81.4 | 56.3 | 93.8 | 79.2 | **85.5** | **72.9** | 62.4 | 77.0 |
| | Same layer connection | **79.0** | **98.0** | **85.6** | **92.8** | 57.5 | **61.3** | **69.7** | **74.8** | **82.4** | **93.1** | **63.1** | **95.2** | **84.8** | **68.0** | **95.3** | 80.7 | 85.5 | 62.8 | **71.3** | **79.6** |
| | Maltiple layer connection | 77.1 | 97.7 | 84.4 | 91.9 | 48.3 | 58.2 | 68.0 | 73.3 | 81.4 | 92.7 | 62.5 | 95.0 | 84.1 | 65.9 | 94.9 | 75.7 | 84.4 | 59.1 | 69.3 | 79.0 |
| DeepLabv3+ | Single | 77.3 | **98.7** | 83.6 | 92.5 | 54.4 | 60.2 | 64.0 | 69.5 | 78.0 | 92.5 | 63.8 | 95.0 | 81.8 | 63.0 | 95.0 | **82.8** | 85.3 | 66.0 | 65.8 | 76.4 |
| | Ensemble | 77.5 | 97.9 | 83.5 | 92.6 | **55.9** | 61.2 | 65.1 | 70.2 | 78.9 | **92.7** | **65.6** | 95.0 | 82.2 | 63.4 | 95.0 | 77.3 | 84.3 | 66.2 | 65.9 | 76.8 |
| | Same layer connection | 77.8 | 98.0 | **84.2** | 92.5 | 53.8 | **60.5** | 65.4 | 70.9 | 79.0 | 92.6 | 64.0 | **95.0** | 82.4 | **65.0** | 94.8 | 73.0 | 85.1 | 68.9 | **68.6** | 77.4 |
| | Maltiple layer connection | **78.3** | 98.0 | 84.1 | **92.8** | 53.2 | 60.3 | **66.2** | **71.4** | **80.3** | 92.6 | 64.2 | 94.8 | **82.8** | 64.3 | **95.1** | 81.0 | **87.4** | **75.3** | 67.5 | 77.4 |

Table2.    Segmentation results on PASCALVOC dataset

| Network | Method | Mean Iou% |
|---|---|---|
| DANet | Single | 80.04 |
| | Ensemble | 80.60 |
| | Same | **81.68** |
| | Maltiple | 78.12 |
| DeepLabv3+ | Single | 78.44 |
| | Ensemble | 78.09 |
| | Same | 79.86 |
| | Maltiple | **80.13** |

## 4  EXPERIMENTS

This section shows experimental results of cooperative learning. To evaluate the proposed method, we carry out comprehensive experiments on the Cityscapes dataset and the PASCAL VOC2012 dataset using DANet and Deeplabv3+.

We explain the details of datasets in section 4.1 We explained implementation details in section 4.2 We show the result of cooperative DANet network in section 4.3 and the result of cooperative DeepLabv3+ network in section 4.4.

### 4.1  Datasets

We used two kinds of datasets. The details of each dataset are described in the following sections.

#### 4.1.1  Cityscapes

This dataset includes images captured by a camera mounted on a car in Germany. All images are 2,048 × 1,024 pixels in which each pixel has high quality 19 class labels. There are 2,979 images in training set, 500 images in validation set. In this study, we randomly cropped the images of 768×768 pixels in train phase. We evaluate our method with validation dataset.

#### 4.1.2  PASCALVOC2012

This dataset includes the various images. There are 10,582 images in training set, 1,449 images in validation set and 1,456 images in test set. These images involve 20 foreground object classes and one background class. In this study, we use validation set to get the best model. We evaluate test set using the model determined by validation phase. In addition, we randomly cropped the images of 513 × 513 pixels from training set in training phase, and we cropped the center region in validation and test phase.

### 4.2  Implementation Details

We implement our method based on Pytorch. The cooperative learning network is evaluated for two kinds of networks; DANet and DeepLabv3+. For the fair comparison, we evaluated single network and the proposed method under the same conditions. We adopt the ResNet101 as backbone in the single network and cooperative network. It is impossible for us to use the batchsize in the original paper because of the memory size of our GPU. Thus, we changed batchsize and the learning late. Due to the above constraints, we compare our method with our implemented results rather than the results in original papers.

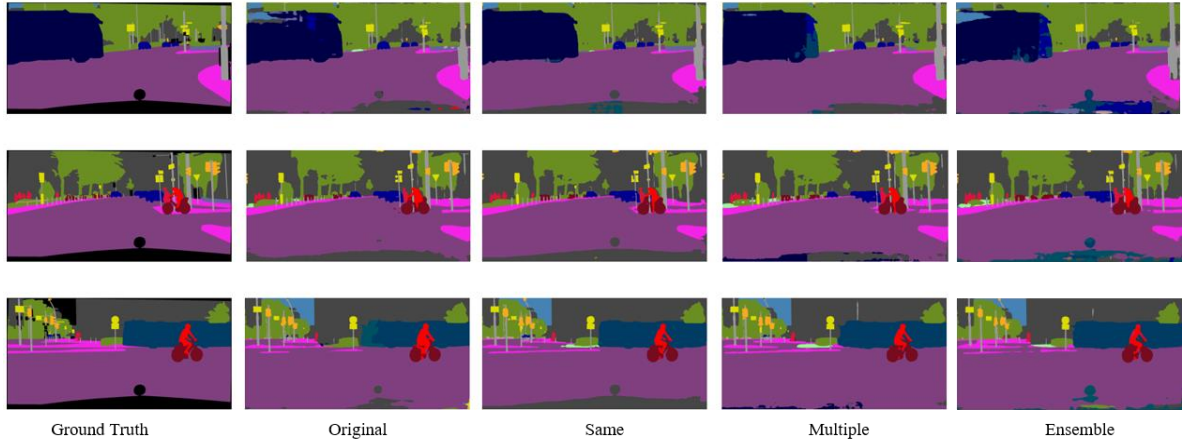

Figure 3. Segmentation results on the Cityscapes dataset (val). The baseline is DANet.

We evaluated four methods. The first one is original single network such as DANet and DeeplabV3+. The second one is the ensemble of two networks. The third one is the proposed method with the same layer connection. The fourth one is the proposed method with multiple layer connection. The accuracy of each single network is used as baseline. We used intersection over union (IoU) and mean IoU (mIoU) as evaluation measures.

## 4.3 Evaluation Using DANet

In this experiment, we used cooperative learning with the same layer connection as shown in Figure 2(b). DANet has two streams; The one stream is used position attention and the other stream is used channel attention. If we use multiple layer connection as shown Figure2 (c), it is considered that various kinds of information by each attention are mixed. It will be bad influence for cooperative learning. Therefore, we made separate cooperative connections on each stream. By using such a connection, we can train cooperative learning of DANet well without mixing the features by two attention mechanisms.

### 4.3.1 Results

This section shows the result of cooperative DANet network. Table1 shows the result on the Cityscapes dataset and Table 2 shows result on the PASCALVOC dataset. According to Table 1, the multiple layer connection improved the accuracy as well as the same layer connection. In comparison with single DANet, cooperative learning achieved 79.02% in mIoU whose improvement is 3.16%. On the other hand, different layer connection also outperforms the baseline by 1.28%. Table 1 shows that almost of all classes improved the accuracy. Especially, the classes with large area such as road, building, bus were improved. This demonstrates that cooperative learning has a benefit to semantic segmentation.

The segmentation results by cooperative networks are shown in Figure 3. We see that our method can recognize the details of class such as road, truck and bus in comparison with baseline. Cooperative learning can use different feature maps for training. This improved the accuracy. However, multiple layer connection gave less performance than the same layer connection. We use the network using attention mechanism in this experiment. Attention emphasizes the important features at each layer for solving the task. Thus, those features may not be effective for sending to the other layers. This is because cooperative learning with multiple layer connection is worse than that with the same layer connection.

Similarly, the same layer connection is highest for PASCALVOC dataset as shown in Table 2 while multiple layer connection gave low accuracy. As described previously, due to negative attention effect in multiple layer connection, it gave lower accuracy for both datasets. Thus, we consider that multiple layer connection does not give good effect when we use CNN with attention.

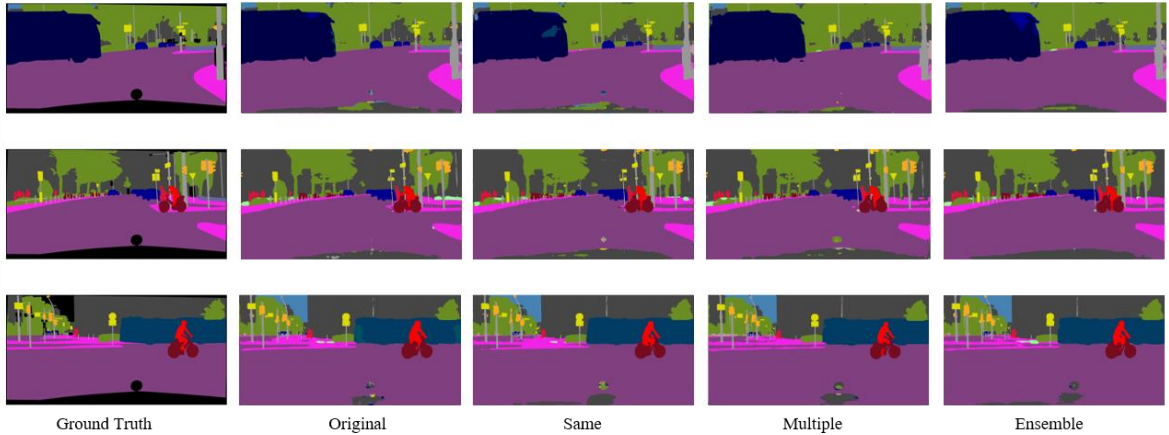

Figure 4. Segmentation results on the Cityscapes dataset (val). The baseline is by Deeplabv3+

## 4.4 Evaluation Using DeepLabv3+

The basic Deeplabv3+ is divided into an encoder part including ASPP and a decoder part. As shown in previous experiments, if separate stream network is used, we change the multiple layer connection because we consider that mixing information in different streams will give bad influence for training. However, this network does not have separate stream or attention mechanism. Thus, we can use basic connection method. In the case of multiple layer connection, we send the feature maps from all top network layers to the bottom network.

### 4.4.1 Results

We show the results by cooperative learning of DeepLabv3+ in Table 1 and 2. Table 1 shows the results on the Cityscapes dataset and Table 2 shows the results on the PASCALVOC dataset. Our method achieved better accuracy than single DeepLabv3+. We compare two connection methods in Table 1. Cooperative network with the same layer connection achieved 77.80% in mIoU which brings 0.5% improvement. Cooperative learning with multiple layer connection improved 1.04% in comparison with the baseline. In addition, we can improve the accuracy in almost of all classes. Our method is able to give a good effect.

Similarly, the accuracy of the proposed method was improved on the PASCAL VOC as shown in Table 2. Each connection gave more than 1% improvement. Cooperative network using Deeplabv3+ was able to improve accuracy on two datasets.

Our method can segment objects well in comparison with standard Deeplabv3+ on Cityscapes dataset in Figure 4. In this experiment, multiple layer connection is better than the same layer connection. This is because DeepLabV3+ is specialized to aggregate the features at multiple scales. Thus, we consider that multiple layer connection can obtain good performance in some feature aggregation networks.

## 4.5 Consideration

From above experiments, we confirmed that the cooperative learning improved the accuracy regardless of the types of baseline networks. For DANet, cooperative learning with the same layer connection gave higher accuracy than that with different layer connection. On the other hand, multiple layer connection gave higher accuracy for DeepLabv3+. From those results, the optimal connection method depends on baseline CNN. DANet used attention module to enhance the feature map. Therefore, the same layer connection can use important features at corresponding layer. On the other hand, different layer connection cannot use those features well at different layers.

In the case of DeepLabv3+, multi-scale information was aggregated using an encoder-decoder structure. Multiple layer connection can provide effective information from multiple layers. As a result, multiple layer connection got high accuracy because many information from different layers help DeepLabv3+ structure to get effective information. We consider that different layer connection is valid using a simple encoder-decoder structure network. If we use an attention module to enhance features map, cooperative learning with the same layer connection is effective.

In experiments, we compared the ensemble of networks as shown in Figure 2(a) with the proposed cooperative network. The proposed method with two kinds of connections is more accurate than the ensemble of two networks. The effectiveness of our cooperative network is demonstrated through two kinds of experiments.

## 5 CONCLUSIONS

In this paper, we proposed a cooperative learning for semantic segmentation that sends the feature maps of top network to the other network. Specifically, we evaluated our methods with two kinds of CNNs and two connection methods. As a result, the effectiveness of our method was demonstrated by experiments on two datasets. Cooperative learning with the same layer connection gave good performance for both networks. However, the improvement of multiple layer connection is small for DANet with attention mechanism. Connection method depends on baseline network structure. In this paper, we use two kinds of connection but many connection methods can be considered. This is a subject for future works.

## ACKNOWLEDGEMENTS

This work is partially supported by MEXT/JSPS KAKENHI Grant Number 18K111382.

## REFERENCES

Krizhevsky, A., Sutskever, I., Hinton, G. E. "ImageNet classification with deep Convolutional neural networks", In Advances in neural information processing systems, pp.1097-1105, (2012)

Szegedy, C., Liu, W., Jia, Y., Sermanet, P., Reed, S., Anguelov, D., Erhan, D., Vanhoucke, V., Rabinovich, A.: Going deeper with convolutions. In: Proceedings of the IEEE conference on Computer Vision and Pattern Recognition. pp. 1–9 (2015)

Redmon, J., Divvala, S., Girshick, R., Farhadi, A.: You only look once:unified, real-time object detection. In: Proceedings of the IEEE Conference on Computer Vision and Pattern Recognition. pp. 779–788 (2016)

Cao, Z., Hidalgo, G., Simon, T., Wei, S.E., Sheikh, Y.: Openpose: realtime multi-person 2d pose estimation using part affinity fields. arXiv preprint arXiv:1812.08008 (2018)

Isola, P., Zhu, J.Y., Zhou, T., Efros, A.A.: Image-to-image translation with conditional adversarial networks. In: Proceedings of the IEEE conference on Computer Vision and Pattern Recognition. pp. 1125–1134 (2017)

Chen, L.C., Collins, M., Zhu, Y., Papandreou, G., Zoph, B., Schroff, F., Adam, H., Shlens, J.: Searching for efficient multi-scale architectures for dense image prediction. In: Advances in Neural Information Processing Systems. pp. 8699–8710 (2018)

Havaei, M., Davy, A., Warde-Farley, D., Biard, A., Courville, A., Bengio, Y., Pal, C., Jodoin, P.M., Larochelle, H.: Brain tumor segmentation with deep neural networks. Medical image analysis 35, 18–31 (2017)

Long, J., Shelhamer, E., Darrell, T.: Fully convolutional networks for semantic segmentation. In: Proceedings of the IEEE Conference on Computer Vision and Pattern Recognition. pp. 3431–3440 (2015)

Ding, H., Jiang, X., Shuai, B., Qun Liu, A., Wang, G.: Context contrasted feature and gated multi-scale aggregation for scene segmentation. In: Proceedings of the IEEE Conference on Computer Vision and Pattern Recognition. pp. 2393–2402 (2018)

Yang, M., Yu, K., Zhang, C., Li, Z., Yang, K.: Denseaspp for semantic segmentation in street scenes. In: Proceedings of the IEEE Conference on Computer Vision and Pattern Recognition. pp. 3684–3692 (2018)

Peng, C., Zhang, X., Yu, G., Luo, G., Sun, J.: Large kernel matters–improve semantic segmentation by global convolutional network. In: Proceedings of the IEEE conference on Computer Cision and Pattern Recognition. pp. 4353–4361 (2017)

Huang, Z., Wang, X., Huang, L., Huang, C., Wei, Y., Liu, W.: Ccnet: Criss-cross attention for semantic segmentation. In: Proceedings of the IEEE International Conference on Computer Vision. pp. 603–612 (2019)

Fu, J., Liu, J., Tian, H., Li, Y., Bao, Y., Fang, Z., Lu, H.: Dual attention network for scene segmentation. In: Proceedings of the IEEE Conference on Computer Vision and Pattern Recognition. pp. 3146–3154 (2019)

Chen, L.C., Zhu, Y., Papandreou, G., Schroff, F., Adam, H.: Encoder-decoder with atrous separable convolution for semantic image segmentation. In: Proceedings of the European Conference on Computer Vision. pp. 801–818 (2018)

Zhang, H., Dana, K., Shi, J., Zhang, Z., Wang, X., Tyagi, A., Agrawal, A.: Context encoding for semantic segmentation. In: Proceedings of the IEEE conference on Computer Vision and Pattern Recognition. pp. 7151–7160 (2018)

Cordts, M., Omran, M., Ramos, S., Rehfeld, T., Enzweiler, M., Benenson, R., Franke, U., Roth, S., Schiele, B.: The cityscapes dataset for semantic urban scene understanding. In: Proceedings of the IEEE conference on Computer Vision and Pattern Recognition. pp. 3213–3223 (2016)

Everingham, M., Van Gool, L., Williams, C.K., Winn, J., Zisserman, A.: The pascal visual object classes (voc) challenge. International journal of computer vision 88(2), 303–338 (2010)

Badrinarayanan, V., Kendall, A., Cipolla, R.: Segnet: A deep convolutional encoder-decoder architecture for image segmentation. IEEE Transactions on Pattern Analysis and Machine Intelligence 39(12), 2481–2495 (2017)

Ronneberger, O., Fischer, P., Brox, T.: U-net: Convolutional networks for biomedical image segmentation. In: International Conference on Medical Image Computing and Computer-Assisted Intervention. pp. 234–241. Springer (2015)

Zhao, H., Shi, J., Qi, X., Wang, X., Jia, J.: Pyramid scene parsing network. In: Proceedings of the IEEE Conference on Computer Vision and Pattern Recognition. pp. 2881–2890 (2017)

Takikawa, T., Acuna, D., Jampani, V., Fidler, S.: Gated-scnn: Gated shape cnns for semantic segmentation. In: Proceedings of the IEEE International Conference on Computer Vision. pp. 5229–5238 (2019)

Song, G., Chai, W. : Collaborative learning for deep neural networks. In Advances in Neural Information Processing Systems. pp. 1832-1841 (2018).

Zheng, H., Xie, L., Ni, T., Zhang, Y., Wang, Y. F., Tian, Q., Elliot, K. F & Yuille, A. L. : Phase Collaborative Network for Two-Phase Medical Image Segmentation. arXiv preprint arXiv:1811.11814. (2018)

Lake, B. M., Salakhutdinov, R., Tenenbaum, J. B.: Human-level concept learning through probabilistic program induction. Science, 350(6266), pp.1332-1338. (2015)